\documentclass[a4paper, journal]{temp/IEEEtran}  

\IEEEoverridecommandlockouts      

\usepackage{multirow}
\usepackage{csquotes}
\usepackage{booktabs}
\usepackage{graphicx}
\usepackage{mathptmx} 
\usepackage{times} 
\usepackage{amsmath} 
\usepackage{amssymb}  

\usepackage{subfigure}
\usepackage{float}
\usepackage{booktabs}
\usepackage{mathalfa}
\usepackage{mathtools}
\usepackage[mathcal]{eucal}
\usepackage[scientific-notation=true]{siunitx}
\usepackage{color}
\usepackage{bm}
\usepackage[dvipsnames]{xcolor}
\usepackage{algorithm}
\usepackage{algpseudocode}
\usepackage{color}
\usepackage{float}
\usepackage[hyperfootnotes=false,hyperfigures=false]{hyperref}

\begin{document}












\title{Adaptive feature recombination and recalibration for semantic segmentation with Fully Convolutional Networks}

\author{S\'{e}rgio Pereira, Adriano Pinto, Joana Amorim, Alexandrine Ribeiro, Victor Alves, and Carlos A. Silva
\thanks{Copyright (c) 2019 IEEE. Personal use of this material is permitted. However, permission to use this material for any other purposes must be obtained from the IEEE by sending a request to pubs-permissions@ieee.org.}
\thanks{S. Pereira, A. Pinto, J. Amorim, A. Ribeiro, and C. A. Silva are with CMEMS-UMinho Research Unit, University of Minho, Guimar\~{a}es, Portugal (e-mail: id5692@alunos.uminho.pt (S. Pereira), csilva@dei.uminho.pt (C. Silva))
       }%
\thanks{S. Pereira, A. Pinto, and V. Alves are with Centro Algoritmi, University of Minho, Braga, Portugal (e-mail: valves@di.uminho.pt (V. Alves))}
}

\maketitle

\begin{abstract}

Fully Convolutional Networks have been achieving remarkable results in image semantic segmentation, while being efficient. Such efficiency results from the capability of segmenting several voxels in a single forward pass. So, there is a direct spatial correspondence between a unit in a feature map and the voxel in the same location. In a convolutional layer, the kernel spans over all channels and extracts information from them. We observe that linear recombination of feature maps by increasing the number of channels followed by compression may enhance their discriminative power. Moreover, not all feature maps have the same relevance for the classes being predicted. In order to learn the inter-channel relationships and recalibrate the channels to suppress the less relevant ones, Squeeze and Excitation blocks were proposed in the context of image classification with Convolutional Neural Networks. However, this is not well adapted for segmentation with Fully Convolutional Networks since they segment several objects simultaneously, hence a feature map may contain relevant information only in some locations. In this paper, we propose recombination of features and a spatially adaptive recalibration block that is adapted for semantic segmentation with Fully Convolutional Networks --- the SegSE block. Feature maps are recalibrated by considering the cross-channel information together with spatial relevance. Experimental results indicate that Recombination and Recalibration improve the results of a competitive baseline, and generalize across three different problems: brain tumor segmentation, stroke penumbra estimation, and ischemic stroke lesion outcome prediction. The obtained results are competitive or outperform the state of the art in the three applications.

\end{abstract}


\section{Introduction}
\bstctlcite{my_bib:BSTcontrol}

Medical image segmentation is often part of medical image analysis pipelines, being crucial for diagnostic, treatment planning, and follow-up \cite{litjens2017survey,menze2015multimodal,maier2017isles}. However, when manually done, segmentation is time demanding and prone to inter- and intra-rater variability \cite{menze2015multimodal,maier2017isles}. Automatic segmentation can mitigate those issues, thus improving the efficiency and quality of medical care, allowing the experts to focus on other tasks \cite{magoulas1999machine}. In recent years, several approaches have been based on machine learning methods. Among them, many rely on Deep Learning using Convolutional Neural Networks (CNNs) \cite{litjens2017survey}.

Convolutional Neural Networks are composed by convolutional layers that learn features directly from the data. Each of these layers consists of a set of learnable kernels that are convolved over the input data to generate a stack of feature maps. These models gathered a lot of interest after the results achieved by Krizhevsky et al. \cite{krizhevsky2012imagenet} in object recognition. Since then, CNN-based approaches have reached state-of-the-art results across many applications in medical image analysis \cite{pereira2016brain,ronneberger2015u,kamnitsas2017efficient}. In recent years, active research is being conducted regarding architectural design in deep neural networks. Several studies focused on enabling training of deeper networks, such as VGGNet \cite{simonyan2014very} that uses small kernels to reduce the number of parameters, or residual learning, elegantly implemented through short skip connections \cite{7780459}. However, other dimensions of CNNs, besides depth, are being studied, such as the cardinality \cite{xie2017aggregated}, or the cross-channels relationships in a stack of feature maps \cite{hu2017squeeze}.

The relationships among the channels may be used to enhance the representational power of the features. For instance, convolutional layers with $1 \times 1$ kernels were used as parametric pooling \cite{lin2013network}, or as bottlenecks in ResNets \cite{7780459}. Interestingly, in the latter, these layers were also used to increase the number of feature maps after the previous bottleneck. Since layers with $1 \times 1$ kernels do not take a neighborhood into account, they work solely by recombining the channels of the feature maps. Differently, Hu et al. \cite{hu2017squeeze} explored the relationships among channels and proposed recalibration of feature maps with Squeeze-and-Excitation (SE) blocks. In this case, the end goal is to adaptively suppress the less relevant feature maps.

Many advances in CNN`s design for image applications are often proposed in the context of object recognition \cite{litjens2017survey}. In this problem, a single class label is inferred for the whole image. Similar approaches can be used for segmentation. For instance, Pereira et al. \cite{pereira2016brain} used a conventional CNN-based method that classifies the central voxel of a patch. So, in this case, it is straightforward to introduce general CNN blocks. However, recently, the more efficient Fully Convolutional Networks (FCNs) \cite{ronneberger2015u,long2015fully} are being preferred for image segmentation over conventional CNNs. In these architectures, the fully-connected layers of CNNs are replaced by convolutional layers. In this way, it is possible to segment a full patch of voxels in just one forward pass. While in conventional CNNs the whole feature maps characterize the class of just one voxel, in FCNs there is a direct correspondence between the features in a given location of the feature maps and the voxel being classified in the same spatial location. Therefore, some architectural designs may not be well suited for FCNs, such as the SE block. The reason is that this block weights whole feature maps, while we may be interested in suppressing regions of the feature maps that are irrelevant when predicting voxels in the same spatial location.

Besides learning channel-wise relationships to enhance the discriminative power of the feature maps, the SE block \cite{hu2017squeeze} may be interpreted as a channel-wise attention mechanism \cite{oktay2018attention}. Different kinds of attention have been recently proposed in the context of medical image segmentation using FCNs. Qin et al. \cite{qin2018autofocus} proposed a scale attention scheme that adaptively chooses the receptive field. This is achieved by processing several parallel branches with convolutional layers with different dilation rates. However, this increases the memory and computational requirements. Oktay et al. \cite{oktay2018attention} utilized attention as a soft mask to enhance the region of interest. However, this approach scales the same regions across all feature maps. Roy et al. \cite{roy} learn attention at the channel and spatial levels, but a single spatial attention map is inferred for all feature maps.

\subsection{Motivation and Contributions}

The cross-channel relationships among feature maps were shown to encode relevant information \cite{hu2017squeeze}. Therefore, in this work, we study the recombination and recalibration of feature maps. Regarding recombination, convolutional layers with $1 \times 1$ kernels were previously used to decrease the number of feature maps \cite{lin2013network,7780459}, but also to increase their number afterwards \cite{7780459} to boost their representational power. We propose recombination, where we linearly expand the number of feature maps before compressing again, allowing the network to learn how to mix the information to generate more discriminative features. In the case of recalibration, the SE block has the desirable property of learning cross-channel relationships and suppressing the less relevant feature maps. However, it recalibrates whole feature maps, which makes it not well adapted for semantic segmentation with FCNs where it may be better to have spatially adaptive recalibration. So, in this work we also propose a recalibration block (the SegSE block) that is able to spatially recalibrate feature maps, while still considering the cross-channel relationships. This can be alternatively interpreted as a spatially adapted attention mechanism for each feature map; unlike \cite{oktay2018attention}, where a single attention map is generated. Related to our work is the block proposed by Roy et al. \cite{roy}. But, channel and spatial recalibration are considered separately, while we learn them jointly. Also, a single spatial recalibration map is estimated for all channels without taking context into account, while we infer channel-specific maps.

An early version of this work was presented at a conference \cite{segse}. This paper extends the previous work with further validation across several datasets and applications. Additionally, we provide more detailed descriptions on the method and discussion, and insights on the working process of the proposed approach. We observe that it indeed learns how to suppress or enhance features accordingly to the structures being segmented. The contributions in this paper can be summarized as follows. 1) We propose feature recombination by means of linear expansion and compression of the number of feature maps. 2) We propose a novel feature recalibration approach for FCNs -- the SegSE block. This block may be interpreted as an attention mechanism for each channel of the stack of feature maps. 3) We validate our methods in several publicly available datasets for brain tumor segmentation, stroke penumbra estimation, and ischemic stroke lesion outcome prediction. Finally, 4) we inspect the attention maps to verify that they enhance specific features, while the original SE block suppresses whole feature maps, including potentially important local features.

The remaining sections of this paper are organized as follows. The proposed methods are presented in Section \ref{sec:methods}. The experimental setup is described in Section \ref{sec:exp_set}. Then, in Section \ref{sec:results}, we present the results and discussion. Finally, the main conclusions are presented in Section \ref{sec:conclusion}.

\section{Methods}
\label{sec:methods}

\begin{figure*}[!htb]
\centering
\subfigure[]{\label{subfig:multi_seg}\includegraphics[width=0.9\textwidth, keepaspectratio]{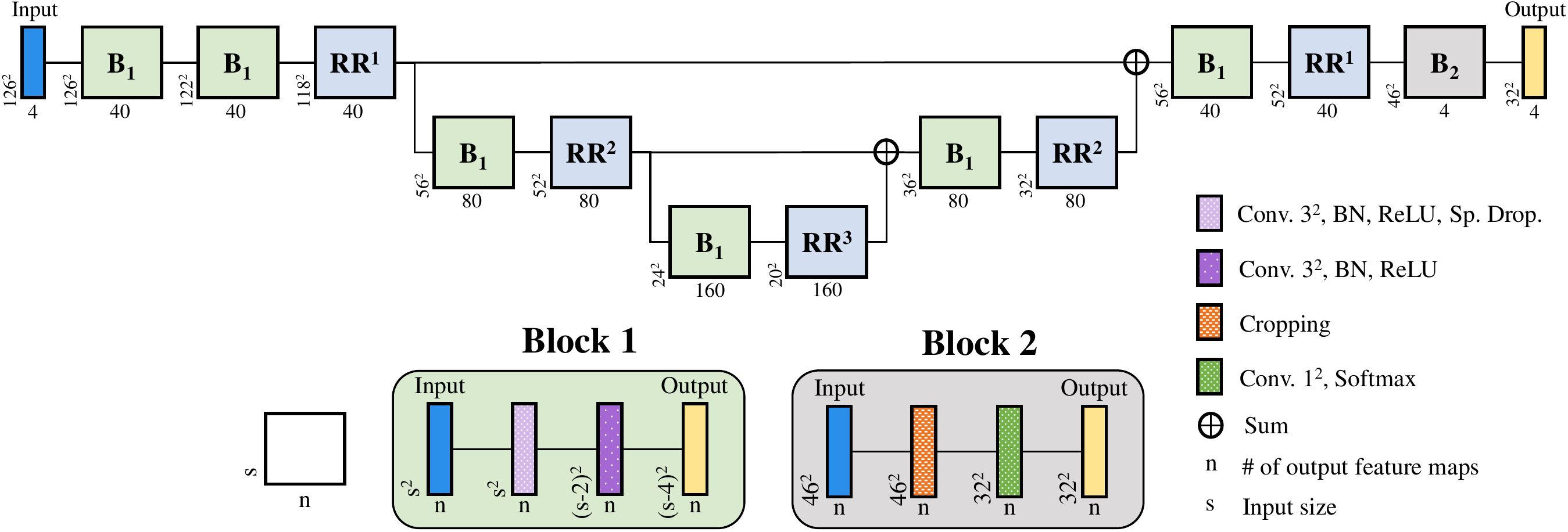}}
\subfigure[]{\label{subfig:recombination}\includegraphics[width=0.25\textwidth, keepaspectratio]{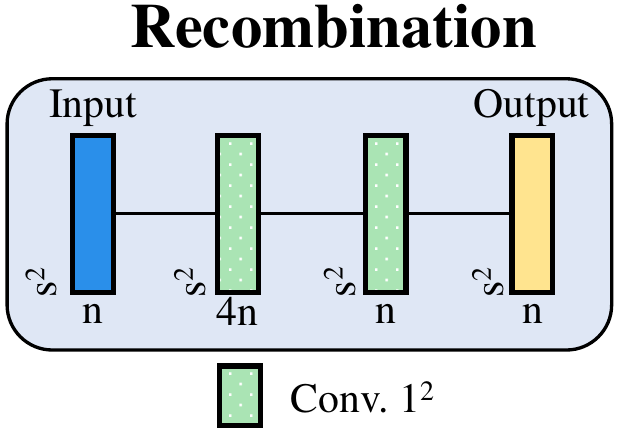}} 
\hfill
\subfigure[]{\label{subfig:recalibration}\includegraphics[width=0.9\textwidth, keepaspectratio]{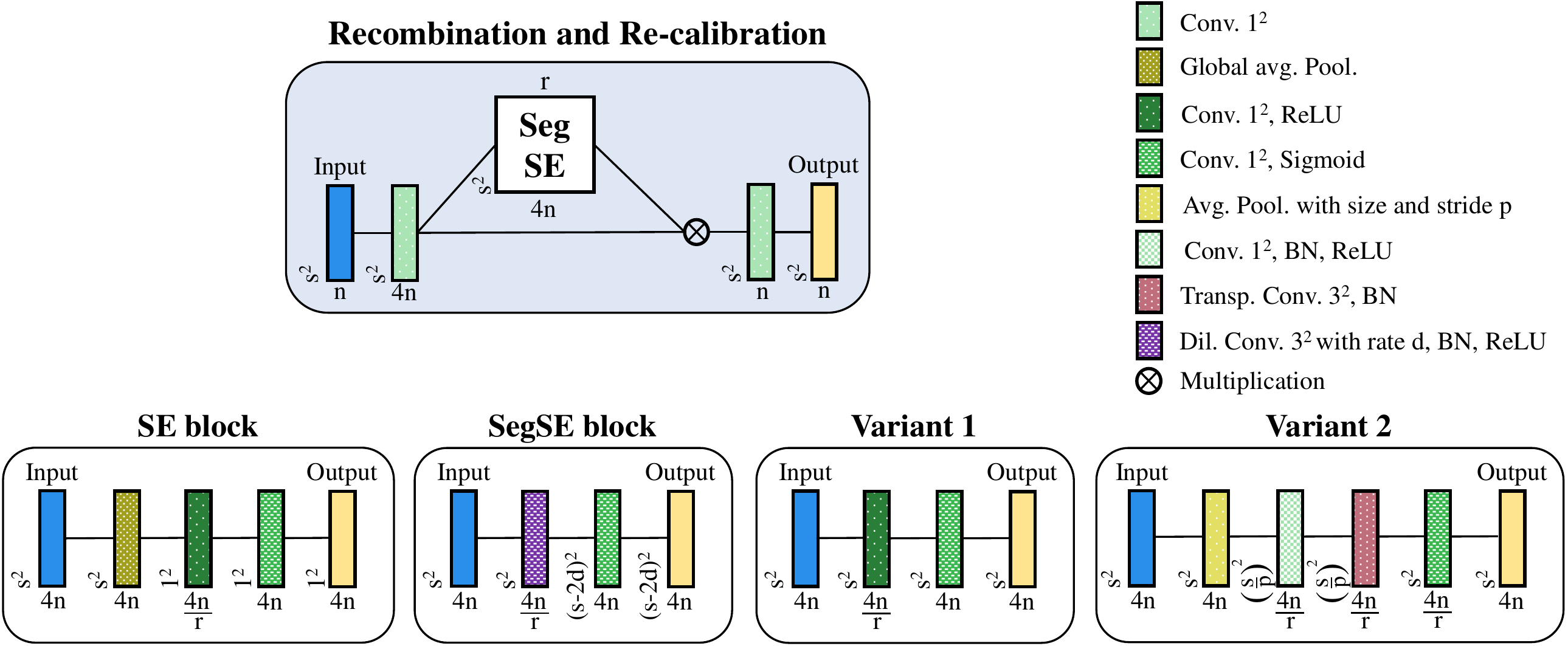}}
\caption{U-Net-inspired FCN and the proposed recombination and recalibration (RR) blocks. Each rectangle represents a layer or a set of layers (blue and yellow rectangles are the input and output, respectively). a) Architecture overview with the RR block. B$_1$ and B$_2$ represent Block 1 and Block 2, respectively. Input sizes correspond to the RR block with SegSE. Down-sampling is obtained by max-pooling. We use nearest neighbor up-sampling to increase the feature maps size, and $1\times 1$ convolutional layers to adjust the number of feature maps, before addition. b) Recombination block. c) RR block, the SE block, the SegSE blocks, and the other variants. BN stands for batch normalization, and Sp. Drop. for spatial dropout.}
\label{fig:senet}
\end{figure*}

In this work, we approach semantic segmentation using 2D FCNs operating over patches. The architecture of our network is inspired in U-Net \cite{ronneberger2015u}, as depicted in Fig. \ref{subfig:multi_seg} (note that rectangles represent a layer or set of layers). This type of FCN is well-established in medical image segmentation tasks \cite{litjens2017survey}. U-Net belongs to a broader class of architectures called encoder-decoder. As input we have an image patch, which may have several stacked channels. The encoder path encompasses different levels of abstraction, which are responsible for learning higher order features. Features computed by higher (deeper) convolutional layers are more abstract. However, these features may lack the fine details that are important for segmentation, which are better captured by the lower layers. Since the feature maps are down-sampled, we need to map the lower resolution feature maps back to the input patch resolution. This is gradually done by up-sampling in the decoder path. As we up-sample feature maps, we sum them with the feature maps of equivalent size of lower layers of the encoder path, through long skip connections. Further convolutional layers fuse the lower and higher level features. Finally, the last layer is a convolutional layer with $1\times 1$ kernels followed by the softmax activation function that infers the probability of each voxel belonging to each class \cite{ronneberger2015u}.

We may think of a layer in a FCN as implementing a transformation function ($\mathrm{F^{Tr}}$) that maps the input feature maps $\bm{X}$ into some output feature maps $\bm{U}$. Therefore, it can be defined as $\mathrm{F^{Tr}}: \bm{X} \rightarrow \bm{U^{Tr}}$, with $\bm{X} \in \mathbb{R}^{H'\times W'\times C'}$ and $\bm{U}^{Tr} \in \mathbb{R}^{H\times W\times C}$, where $H$ and $W$ represent the height and width of the transformed feature maps, respectively. $H'$ and $W'$ represent the same measures but in relation to the input feature maps $\bm{X}$. $C$ and $C'$ are the number of feature maps, such that $\bm{X}=\big[\bm{x}_1, \bm{x}_2, \cdots, \bm{x}_{C'}\big]$, and $\bm{U^{Tr}}=\big[\bm{u}_1, \bm{u}_2, \cdots, \bm{u}_{C}\big]$. So, a convolutional layer defines a function $\mathrm{F^{conv}}$, such that $U^{conv}=\mathrm{F^{conv}}\left( \ \cdot \ ; \ k, \ d, \ n \right)$, where $k$, $d$, and $n$ represent the kernel size, the dilation rate, and the number of kernels, respectively. In this case, each output channel $c$ is computed as $u_c = \bm{v}^c * \bm{X} = \sum_{l=1}^{C'} \bm{v}_l^c * \bm{X}_l$, where $\bm{V} = \big[\bm{v}^1, \bm{v}^2, \cdots, \bm{v}^C\big]$ is the set of learnable kernels (the bias term is not represented for the sake of simplicity), and $*$ is the convolution operation symbol.

\subsection{Recombination}
\label{sec:recomb}

In recombination we are interested in increasing the representational power of the features by mixing them linearly. To this end, we employ convolutional layers with $1\times 1$ kernels. First, we expand the number of features maps ($\mathrm{F^{exp}}$), before compressing again to the original number ($\mathrm{F^{comp}}$). So, recombination is defined as $\mathrm{F^{recomb}}: \bm{X} \rightarrow \bm{U}^{recomb}$, with $\bm{U}^{recomb} \in \mathbb{R}^{H' \times W' \times C'}$. This translates into the following operation:

\begin{equation}
	\begin{multlined}
		\mathrm{F^{recomb}}\left( \bm{X} \right) = \mathrm{F^{comp}}\left( \mathrm{F^{exp}} \left( \bm{X}; \ mC' \right); \ C' \right) = \\ \mathrm{F^{conv}}\left( \mathrm{F^{conv}} \left( \bm{X}; \ 1, \ 1, \ mC' \right); \ 1, \ 1, \ C' \right),
	\end{multlined}
\end{equation}

\noindent where $m$ is the expansion factor. These operations are depicted in Fig. \ref{subfig:recombination}.

\subsection{Recalibration}

\begin{figure}[b]
\centering
\subfigure[]{\label{subfig:se_diagram}\includegraphics[width=0.97\columnwidth, keepaspectratio]{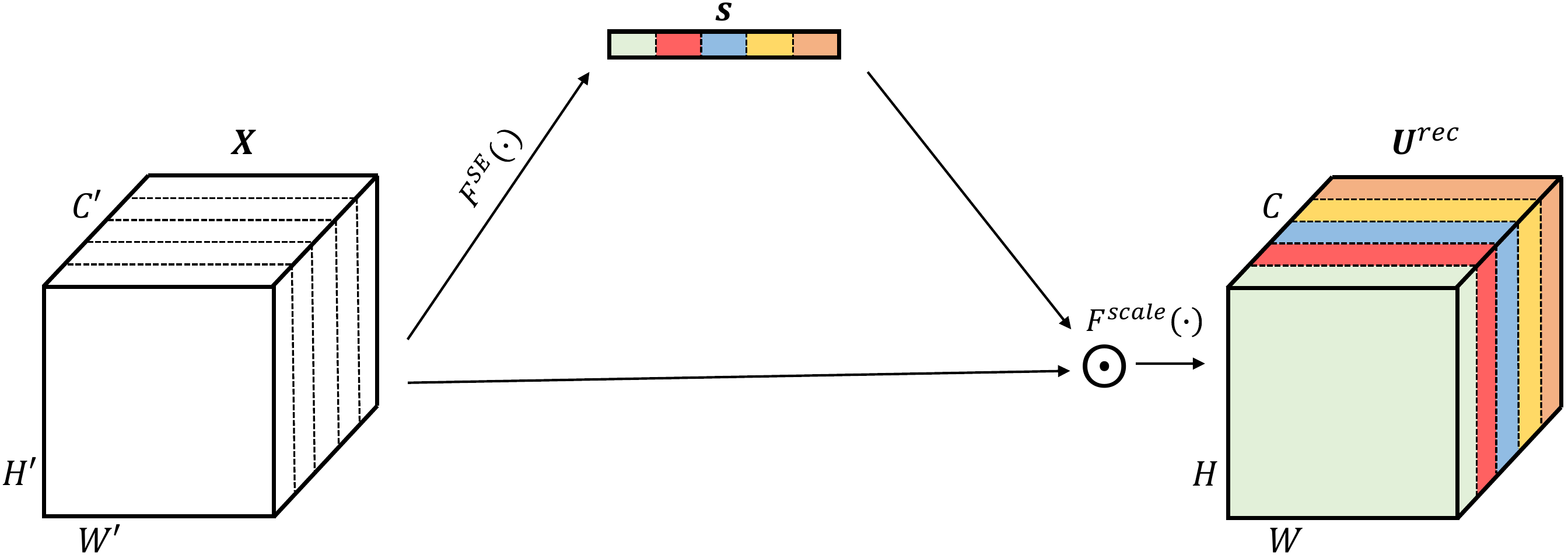}}
\subfigure[]{\label{subfig:segse_diagram}\includegraphics[width=0.97\columnwidth, keepaspectratio]{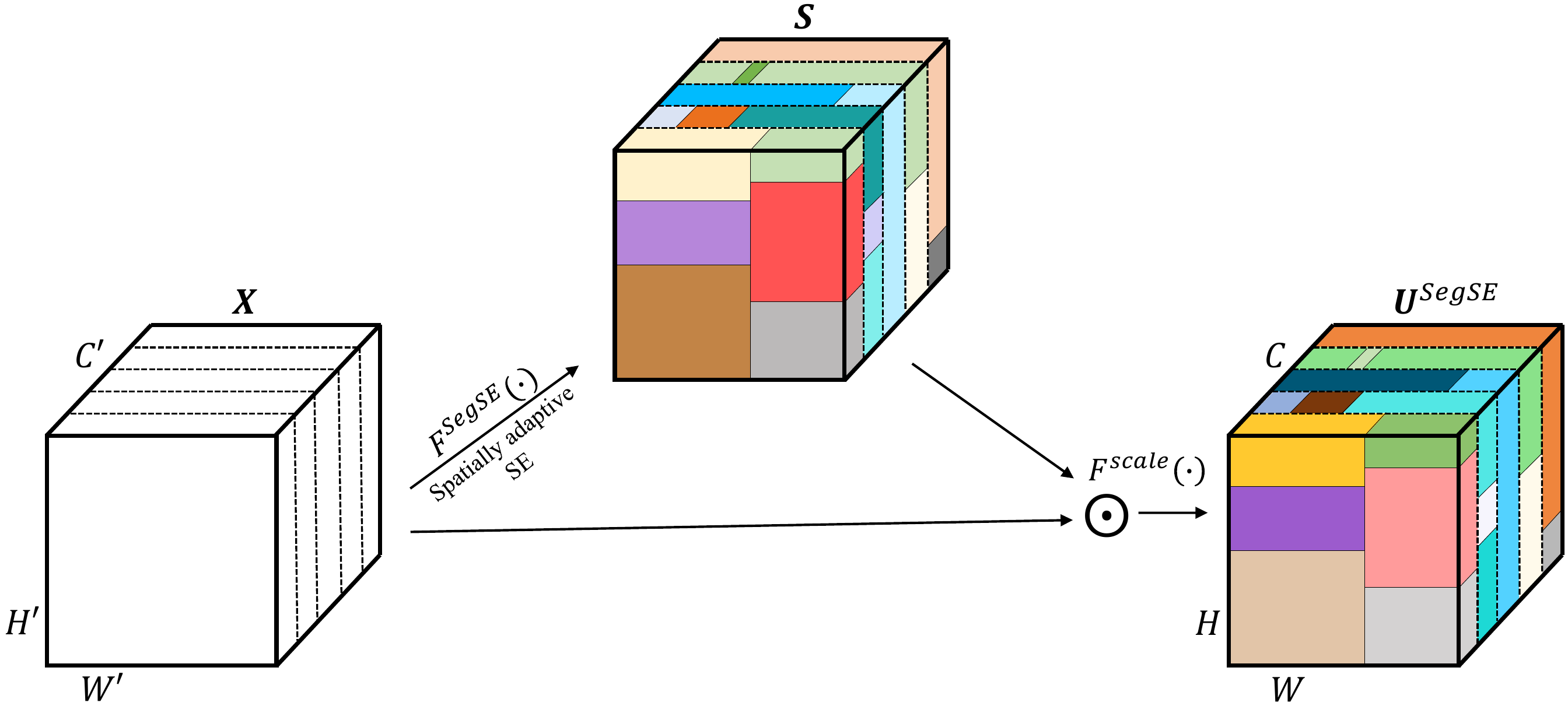}} 
\caption{Depiction of recalibration of feature maps: a) Channel Squeeze-and-Excitation, and b) spatially adaptive Squeeze-and-Excitation.}
\label{fig:recal_diagram}
\end{figure}

Recalibration consists in learning the relationship among the feature maps of a layer, and suppressing the less relevant features. Originally, Hu et al. \cite{hu2017squeeze} proposed to recalibrate whole feature maps, which we refer as channel SE blocks (c.f. Fig. \ref{subfig:se_diagram}). However, in this work, we propose spatially adaptive recalibration (c.f. Fig. \ref{subfig:segse_diagram}).

\subsubsection{Channels Squeeze-and-Excitation}

Let us consider a function $\mathrm{F^{rec}}: \bm{X} \rightarrow \bm{U}^{rec}$, with $\bm{U}^{rec} \in \mathbb{R}^{H \times W \times C}$, that transforms the input feature maps into their recalibrated form. This is achieved by two sequential operations: squeeze and excitation \cite{hu2017squeeze}. The spatial squeeze operation consists in summarizing each channel into a scalar descriptor, such that we obtain a descriptor vector $\bm{z} = \big[ z_1, z_2, \cdots , z_{C'} \big]$, with $\bm{z} \in \mathbb{R}^{1 \times 1 \times C'}$. This operation can be performed by global average pooling, where a given $c$ channel $\bm{x}_c$ is squeezed as,

\begin{equation}
z_c = F^{sq}\left( \bm{x}_c \right) = \frac{1}{H' \times W'} \sum_{i=1}^{H'} \sum_{j=1}^{W'} \bm{x}_c \left( i, j \right).
\end{equation}

In the excitation operation, the network learns the dependencies among the channels in order to adaptively estimate the excitation or scaling factors $\bm{s}$. This can be understood as a gating mechanism that provides channel-wise attention. To this end, two fully-connected layers\footnote{Equivalently implemented as convolutional layers with $1 \times 1$ kernels.} are employed, where the first one acts as bottleneck and the second one restores the dimension of the vector. These operations are described as,

\begin{equation}
\bm{s} = F^{exc} \left( \bm{z} \right) = \sigma\left( \bm{W}_2 \delta \left( \bm{W}_1 \bm{z} \right) \right), 
\end{equation}

\noindent with $\bm{W}_1 \in \mathbb{R}^{C' \times \frac{C'}{r}}$, and $\bm{W}_2 \in \mathbb{R}^{\frac{C'}{r} \times C'}$. $\delta$ denotes the ReLU activation function, $\sigma$ denotes the sigmoid function, and $r$ denotes the reduction factor. Therefore, we can define the function $\mathrm{F^{SE}} \left( \bm{X} \right) = \mathrm{F^{exc}} \left( \mathrm{F^{sq}} \left( \bm{X} \right) \right)$. These operations are visually described in the SE block of Fig. \ref{subfig:recalibration}.

Finally, the recalibration, or scaling, of the feature maps is done by simple multiplication of the original channels by the corresponding recalibration factor. So, it may be defined as $\mathrm{F^{rec}} \left( \bm{X} \right) = \mathrm{F^{scale}} \left( \bm{X}, \ \mathrm{F^{SE}} \left( \bm{X} \right) \right)$. In this way, the recalibration of a feature map $c$ is defined as,

\begin{equation}
\bm{u}_c^{rec} = \bm{x}_c \cdot s_c.
\end{equation}

A visual depiction of channel SE can be found in Fig. \ref{subfig:se_diagram}.

\subsubsection{Spatially adaptive Squeeze-and-Excitation}

In semantic segmentation with FCNs there is a spatial correspondence between the units in the features maps and the pixels/voxels being segmented in the same locations. Hence, a given feature map may have relevant features for some voxels, whereas in other locations it may not be so important. Therefore, due to its global squeeze operation, channel SE blocks may end up suppressing whole feature maps that may contain important regions. For this reason, we argue that spatially adaptive SE blocks, as depicted in Fig. \ref{subfig:segse_diagram}, are more appropriate for semantic segmentation, at least with FCNs.

In this subsection, we start by presenting our simultaneous spatial and channel Squeeze-and-Excitation proposal. Next, we describe two variants that were studied in this work.

\paragraph{Spatially adaptive Squeeze-and-Excitation for segmentation --- SegSE}

The proposed block can be observed in Fig \ref{subfig:recalibration} --- SegSE block. In order to preserve the spatial structure and correspondence of the feature maps, we first replace the squeeze operation through global average pooling in the SE block by a convolutional layer with $3 \times 3$ kernels. The motivation for the squeeze operation is to aggregate contextual information, which in this case is captured by the kernel operating over neighboring voxels. Hence, we employ convolutional layers with dilated kernels \cite{dilated} to capture a larger context than simple $3 \times 3$ kernels, but without increasing the number of parameters over those kernels. In this way, instead of having a descriptor vector $\bm{z}$, as in the case of the channels SE block, we obtain feature maps as,

\begin{equation}
\bm{Z}^{segSE} = \gamma \left( \mathrm{F^{conv}} \left( \bm{X}; \ k^{segSE}, \ d, \ n^{segSE} \right) \right),
\end{equation}

\noindent where $n^{segSE} = \frac{C'}{r}$, $k^{segSE} = 3$, and $\gamma$ represents batch normalization followed by the ReLU activation function. The dilation factor $d$ may be chosen according to the scale of the layer it is operating. Layers that already take a large field of view and context into account may require less dilation. In general, the field of view increases, for instance, with the number of convolutional, or pooling layers. After these layers, each unit of the feature maps represents a larger region of the input space.

Having $\bm{Z}^{segSE}$, the feature maps with recalibration factors are obtained by a convolutional layer with $1 \times 1$ kernels, followed by the sigmoid activation function, as

\begin{equation}
\bm{S} = \sigma \left( \mathrm{F^{conv}} \left( \bm{Z}^{segSE}; \ k, \ d, \ n \right) \right),
\end{equation}

\noindent with $k=1$, $d=1$, and $n=C'$. Therefore, we combine the squeeze and excitation procedures, since the convolutional layer with dilation also includes the bottleneck by decreasing the number of feature maps. Finally, the recalibrated feature maps are obtained by element-wise multiplication ($\odot$) of the input with $\bm{S}$. So, having a feature map $c$, it is recalibrated as

\begin{equation}
\bm{u}_c^{segSE} = \bm{x}_c  \odot \bm{s}_c,
\label{eq:recal}
\end{equation}

In this way, our SegSE block may be described by the function $\mathrm{F^{segSE}} : \bm{X} \rightarrow \bm{U}^{segSE}$.

\paragraph{Variant 1 - No context}

A simpler approach consists in removing the global average pooling in the SE block, and replacing the fully-connected layers by convolutional layers with $1 \times 1$ kernels (Fig. \ref{subfig:recalibration} --- variant 1). Taking the previous description into account, this would translate into setting $k^{segSE} = 1$ and $d=1$. The issue with this approach is that contextual information is not considered, which we previously obtained through $3 \times 3$ convolutional kernels with dilation.

\paragraph{Variant 2 - Pooling-based context}

Contextual information may be obtained by average pooling. Contrasting with global pooling, this acts over a kernel instead of the complete channel of the feature maps. This can be implemented through a pooling layer defined as $\mathrm{F^{avg. pool.}} : \bm{X} \rightarrow \bm{U}^{avg. pool.}$, such that $\bm{U}^{avg. pool.} = \mathrm{F^{avg. pool.}} \left( \bm{X}; \ p \right)$, where $p$ is the size of the pooling kernel and stride, which we assume to be an even number. Therefore, we obtain $\bm{U}^{avg. pool.} \in \mathbb{R}^{\frac{H'}{p} \times \frac{W'}{p} \times C'}$. Then, a convolutional layer with $1 \times 1$ kernels combines the feature maps and learns their relationship.



Average pooling results in a down-sampled feature map. Hence, it is necessary to restore it to the original shape. We accomplish this through transposed convolutional layers \cite{dumoulin2016guide}. We employ these layers to double the size of the feature maps each time. So, if $p > 2$, it is necessary to employ more than one layer of transposed convolution. Therefore, the result of this block is $U^{convTblock} = \mathrm{F^{convTblock}} \left( \bm{X}; \ k_T = 3, \ s_T = 2, n_T=\frac{C'}{r}, \ l_T = \frac{p}{2} \right)$, where $k_T$ is the kernel size, $s_T$ is the stride, $n_T$ is the number of convolutional kernels, and $l_T$ is the number of transposed convolution layers. Each $l_T$ operation consists of $ \theta \left( \mathrm{F^{convT}} \left( \cdot ; k_T, \ s_T, \ n_T \right) \right) $, with $\theta$ being batch normalization.

Having described these operations, this \textit{variant 2} is defined by the following procedures (and in Fig. \ref{subfig:recalibration} --- variant 2),

\begin{equation}
\bm{Z}^{var 2} = \gamma \left( \mathrm{F^{conv}} \left( \mathrm{F^{avg. pool}} \left( \bm{X}; p \right); 1, \ 1, \frac{C'}{r} \right) \right),
\end{equation}

\begin{equation}
	\begin{multlined}
		\bm{S}^{var 2} = \sigma \left( \mathrm{F^{conv}} \left( \bm{Z}^{var 2}; \ k_T, \ s_T, \ n_T, \ l_T \right); 1 , 1 , C' \right),
	\end{multlined}
\end{equation}

\noindent where, $k_T = 3$, $s_T = 2$, $n_T=\frac{C'}{r}$, and $l_T = \frac{p}{2}$.

Finally, recalibration is obtained in a similar way as in equation \ref{eq:recal}.

\subsection{Recombination and Recalibration}

We combine both recombination and recalibration (RR) into the same block, as illustrated in Fig. \ref{subfig:recalibration}. As in recombination, we first expand the number of feature maps, then we recalibrate them, and, finally, we compress the number of feature maps into the original number. This can be defined as,

\begin{equation}
\bm{U}^{RR} = \mathrm{F^{RR}} \left( \bm{X} \right) = \mathrm{F^{comp}} \left( \mathrm{F^{rec}} \left( \mathrm{F^{exp}} \left( \bm{X}; \ mC' \right); \ k, \ d, \ n \right); \ C' \right)
\end{equation}

\section{Experimental Setup}
\label{sec:exp_set}

We evaluate the proposed blocks in three medical applications: brain tumor segmentation, penumbra estimation in acute ischemic stroke, and ischemic stroke lesion outcome prediction. Although all the applications use Magnetic Resonance Imaging (MRI), they are quite diverse in terms of acquisitions. Brain tumor segmentation only takes structural MRI sequences into account. Penumbra estimation has a mixture of low resolution structural, diffusion, and perfusion MRI acquisitions. Finally, in ischemic stroke lesion outcome prediction we deal with perfusion and diffusion acquisitions only. Besides the differences in imaging data, these applications are distinct, too. Brain tumor represents a multi-class classification problem, while the others are binary problems. Additionally, in ischemic stroke lesion outcome prediction we are interested in predicting the status of a stroke lesion three months after the image acquisition and intervention. This differs from a pure segmentation problem, where we segment the objects that are visible at the moment.

The proposed blocks were studied in the brain tumor segmentation task. Then, the best hyperparameters and network architecture were evaluated in the other applications for further validation.

Some hyperparameters were kept constant across experiments. The expansion factor $m$ for recombination  was set to 4. In the case of recalibration, the dilation factors $d$ of our SegSE blocks were defined according to the scale of the feature maps (c.f. Fig. \ref{subfig:multi_seg}) as $ \{ RR^1 , RR^2 , RR^3 \} = \{ 3, 2, 1 \}$. Similarly, the kernel sizes and strides $p$ of the average pooling of \textit{Variant 2} were defined as $ \{ RR^1 , RR^2 , RR^3 \} = \{ 4, 2, 2 \}$. Finally, we set the reduction factor $r$ to $10$. These hyperparameters were tuned using the validation set. During training, the cross entropy loss function was minimized using the Adam optimizer \cite{kingma2014adam} with learning rate of \num{5e-5}. Furthermore, we employed spatial dropout with probability of 0.05, and weight decay of \num{1e-6}. Artificial data augmentation consisted of random sagittal flipping and random rotations of $\{0^\circ, 90^\circ, 180^\circ, 270^\circ \}$. Patches were extracted from the axial plane of the MRI images. The FCNs were implemented using Keras with Theano backend. The proposed blocks are available in an online repository\footnote{\url{https://github.com/sergiormpereira/rr_segse}}.

\subsection{Brain tumor segmentation}
\label{sec:brain_tumor_train}

For the task of brain tumor segmentation we used the Brain Tumor Segmentation Challenge (BRATS) 2017 and 2013 datasets \cite{menze2015multimodal,bakas2017advancing}. BRATS 2017 has two publicly available sets: Training (285 subjects) and Leaderboard (46 subjects). BRATS 2013 encompasses three sets: Training (30 subjects), Leaderboard (25 subjects), and Challenge (10 subjects). For each subject, there are four MRI sequences available: T1-weighted (T1), post-contrast T1-weighted (T1c), T2-weighted (T2), and Fluid-Attenuated Inversion Recovery (FLAIR). All images are already interpolated to $1 \ mm$ isotropic resolution, skull stripped, and aligned. We further pre-processed them by correcting the bias field \cite{tustison2010n4itk}, and standardizing the intensity histogram of each MRI sequence \cite{nyul2000new}. Only the Training sets contain manual segmentations publicly available. In BRATS 2017 it distinguishes three tumor regions: edema, necrotic/non-enhancing tumor core, and enhancing tumor. In BRATS 2013 the manual segmentations have necrosis and non-enhancing tumor separately, although we fuse these labels to be similar to BRATS 2017. Hence, the last layer of the FCN has 4 feature maps, three for tumor classes and one for normal tissue. Evaluation is performed for the whole tumor (all regions combined), tumor core (all, excluding edema), and enhancing tumor. Since annotations are not publicly available for 2017 Leaderboard, 2013 Leaderboard, and 2013 Challenge, metrics are computed by the CBICA IPP\footnote{\url{https://ipp.cbica.upenn.edu/}} and SMIR\footnote{\url{https://www.smir.ch/BRATS/Start2013}} online platforms. The development of the RR block was conducted in the larger BRATS 2017 Training set, which was randomly divided into training (60\%), validation (20\%), and test (20\%); the identification of the subjects in each set is also available in the online repository. All the hyperparameters were found using the validation set, before evaluation in the test set. Afterwards, the test set was added to the training and the FCN was fine-tuned before evaluation in the Leaderboard set. However, networks tested in BRATS 2013 were trained in the 2013 Training set.

Given the problem of data imbalance between the tumor tissues and normal tissue, and following recent developments in brain tumor segmentation \cite{pereira2017hierarchical,pinto2018hierarchical,wang2017automatic}, we employ a hierarchical FCN-based brain tumor segmentation approach, as described in \cite{segse}. First, we detect the whole tumor as a binary segmentation problem. Then, we use this information to identify the multi-class segmentation object representing the tumor tissues in the region of interest. The binary brain tumor segmentation FCN is 3D and possesses a large field of view. These two properties are important to reduce the number of false positive detections. The architecture is depicted in Fig. \ref{fig:grade}, and the pipeline for brain tumor segmentation can be found in the Supplementary Material. Note that the 3D binary whole tumor detection network does not contain the RR block and is kept constant across all experiments. Therefore, the variants of the RR block in the more challenging multi-class FCN are the only source of variation in results, allowing us to better compare them and evaluate the benefits of our proposal.

\begin{figure*}[!hbt]
\centering
\includegraphics[width=0.9\textwidth, keepaspectratio]{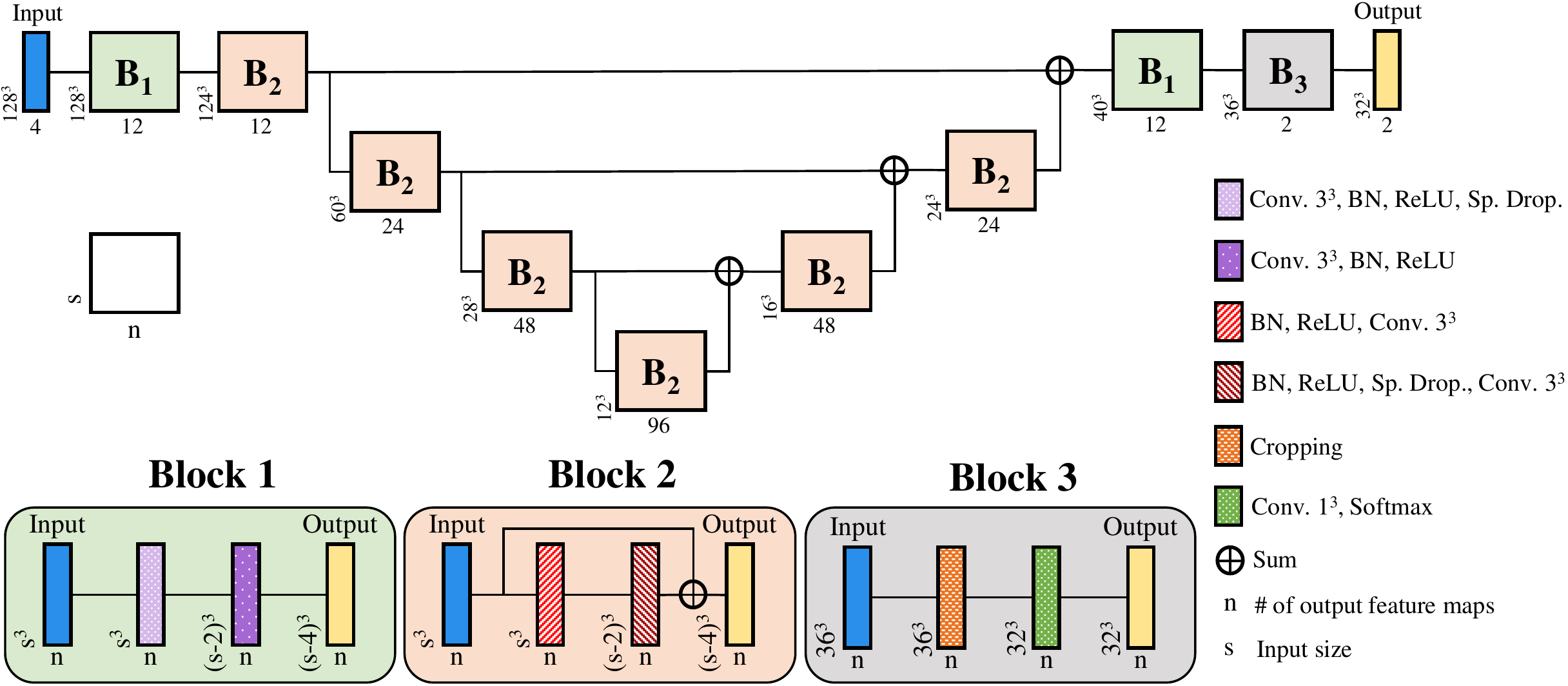}
\caption{Architecture of the binary whole tumor detection FCN. Each rectangle represents a layer or a set of layers (blue and yellow rectangles are the input and output, respectively). Downsampling is obtained by max-pooling. We use nearest neighbor up-sampling to increase the size of the feature maps, and $1\times 1\times 1$ convolutional layers to adjust the number of feature maps before addition. BN stands for batch normalization, and Sp. Drop. for spatial dropout.}
\label{fig:grade}
\end{figure*}

\subsection{Stroke penumbra estimation}

In the case of penumbra estimation in acute ischemic stroke, we used the Stroke Perfusion Estimation (SPES) dataset of the MICCAI Ischemic Stroke Lesion Segmentation (ISLES) Challenge \cite{maier2017isles}. This is a binary classification problem, so, the last layer of the FCN outputs two feature maps. The dataset contains two sets: Training and Challenge. Training has 30 subjects with publicly available annotations. The Challenge set does not have publicly available annotations of its 20 patients, so, evaluation is performed by an online platform\footnote{\url{https://www.smir.ch/ISLES/Start2015}}. Each patient contains 7 sequences: T1c, T2, Diffusion Weighted Imaging (DWI), Cerebral Blood Flow (CBF), Cerebral Blood Volume (CBV), Time-to-Peak (TTP), and Time-to-Max (Tmax). The images are already registered to the T1c, interpolated to a resolution of $2 \times 2 \times 2$ mm resolution, and skull stripped. Further pre-processing included the bias field correction \cite{tustison2010n4itk} and histogram standardization \cite{nyul2000new} of the structural sequences, the clipping of the Tmax values over 60 ($Tmax > 6$ s), and linear scaling of all sequences to the $\big[0, \ 255 \big]$ intensity range.

\subsection{Ischemic stroke lesion outcome prediction}

In this experiment, we used the dataset from ISLES 2017 Challenge \cite{winzeck2018isles}. Similarly to penumbra estimation, this is a binary problem. There are two sets available: Training (43 patients), and Challenge (32 patients). While the former has publicly available annotations, in the latter the evaluation is conducted by an online platform\footnote{\url{https://www.smir.ch/ISLES/Start2017}}. For each patient, there are available 5 perfusion maps (CBV, CBF, Mean Transit Time (MTT), TTP, and Tmax), and 1 diffusion map (ADC). The images were already aligned and skull stripped. Further pre-processing was similar to \cite{pinto2018enhancing}, and consisted in resizing the images to $256 \times 256 \times 32$, clipping of the Tmax values to $[0, 20s]$ and the ADC values to $[0, 2600] \times 10^{-6} \ mm^2/s$, and linear scaling to $[0, 255]$. Finally, and following \cite{pinto2018enhancing}, we employed the Dice loss function during training.

\subsection{Evaluation metrics}

For quantitative evaluation we follow the metrics used in each challenge associated with each dataset. Therefore, for BRATS 2017 Leaderboard we use the Dice Coefficient (DC) and the 95\textsuperscript{th} percentile of the Hausdorff Distance (HD$_{95}$). However, BRATS 2013 employs the DC together with the Sensitivity and Positive Predictive Value (PPV) metrics. In the case of SPES, we report DC and Average Symmetric Surface Distance (ASSD). Regarding ISLES 2017, results are reported in terms of DC, PPV, and Sensitivity. The DC is a measure of overlap, but it is sensitive to the size of the lesions, and does not provide information regarding over- and under-segmentation. Nonetheless, such behavior can be inferred from Sensitivity and PPV. The distance metrics provide insights regarding the correctness of the contour of the segmentations \cite{menze2015multimodal,maier2017isles,winzeck2018isles}.

\section{Results and Discussion}
\label{sec:results}

In this section we evaluate the proposed recombination and recalibration blocks, and we observe experimentally the improvements obtained by the SegSE block. Then, we evaluate the baseline model and the best model in the independent sets of BRATS, SPES, and ISLES 2017. Finally, we compare with the state of the art.

\subsection{Evaluation of Recombination and Recalibration}

\begin{figure}[t]
\centering
\includegraphics[width=1.0\columnwidth, keepaspectratio]{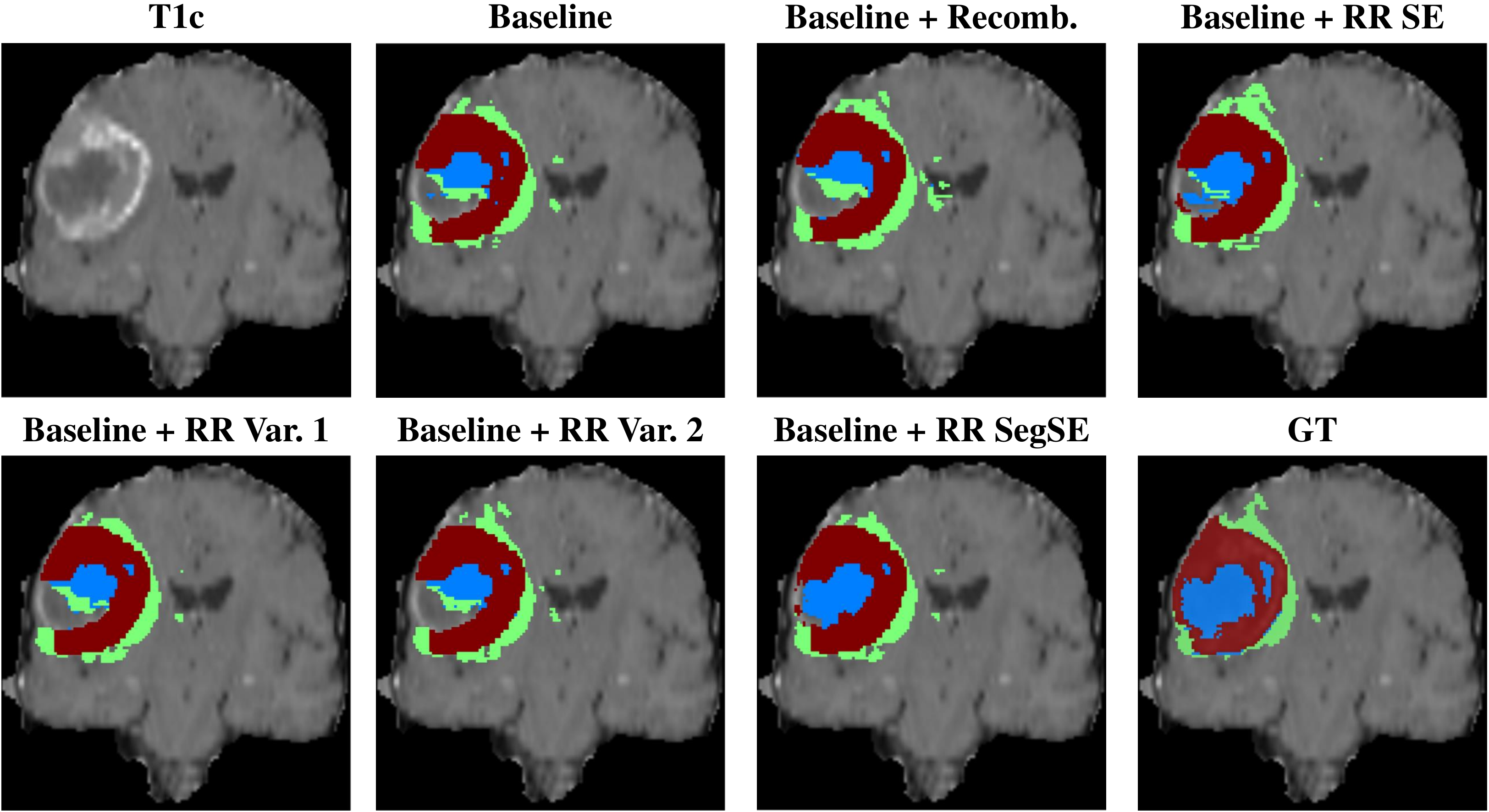}
\caption{Segmentation examples obtained by the baseline architecture and each of the evaluated blocks. The meaning of the colors in the segmentation is: blue --- non-enhancing tumor core, red --- enhancing tumor, and green --- edema. The subject can be found in BRATS 2017 with ID Brats17\_TCIA\_430\_1.}
\label{fig:seg}
\end{figure}

We evaluate the effect of the variants of recombination and recalibration of feature maps in the brain tumor segmentation task. To that end, we use the controlled test set that consists of 20\% of the BRATS 2017 Training set. Table \ref {tab:res_test} shows the quantitative results, while qualitative results can be found in Fig. \ref{fig:seg} as segmentation examples. So, we start from a competitive baseline (as will be shown in Sections \ref{sec:brats}, \ref{sec:spes}, and \ref{sec:isles}) consisting of a FCN similar to the one depicted in Fig. \ref{subfig:multi_seg}, but where the RR block is absent. Then, we incrementally evaluate the proposed blocks. These results can be found in Table \ref{tab:res_test}. It is possible to observe that the addition of recombination (Subsection \ref{sec:recomb}) through linear expansion followed by compression, leads to better DC and Sensitivity in all the tumor regions. This is especially expressive in the core region, where it achieves the highest DC. However, the HD$_{95}$ increased, which may be due to over-segmentation, since the sensitivity scores are higher than the ones obtained by the baseline.

\begin{table*}[!hbt]
\centering
\caption{Results obtained by the baseline and the proposed blocks in the Test set (20\% of BRATS 2017 Training). We evaluate the effect of recombination (Recomb.), and the RR with the original SE block, the proposed SegSE block, and each of the variants. Bold results show the best score in each tumor region.}
\resizebox{1.8\columnwidth}{!}{
\begin{tabular}{lcccccccccccc}
\specialrule{.2em}{.1em}{.1em}
 &  \multicolumn{3}{c}{DC} & \multicolumn{3}{c}{PPV} &  \multicolumn{3}{c}{Sensitivity}  & \multicolumn{3}{c}{HD$_{95}$}  \\ \cmidrule(l){2-4} \cmidrule(l){5-7} \cmidrule(l){8-10} \cmidrule(l){11-13}
\textbf{Method} & Whole & Core & Enh. & Whole & Core & Enh. & Whole & Core & Enh. & Whole & Core & Enh. \\ \midrule
Baseline & 0.857 & 0.739 & 0.682 & 0.898 & 0.832 & 0.706 & 0.830 & 0.733 & 0.704 & 8.645 & 10.761 & 6.672 \\ 
Wide Baseline & 0.852 & 0.751 & 0.678 & 0.906 & 0.855 & 0.706 & 0.818 & 0.729 & 0.698 & 9.049 & 10.647 & 7.065 \\ 
\textbf{Baseline + Recomb.} & 0.865 & \textbf{0.769} & 0.687 & 0.885 & 0.832 & 0.706 & \textbf{0.859} & 0.768 & 0.705 & 9.720 & 11.453 & 7.790 \\ \midrule
Baseline + RR SE & 0.859 & 0.756 & 0.672 & 0.898 & 0.793 & 0.717 & 0.836 & 0.785 & 0.675 & 8.939 & 13.306 & 7.319 \\ 
Baseline + RR Var. 1 & 0.855 & 0.753 & 0.686 & 0.902 & 0.816 & 0.705 & 0.825 & 0.766 & 0.707 & 8.698 & 12.177 & 7.294 \\ 
Baseline + RR Var. 2 & 0.855 & 0.757 & 0.685 & \textbf{0.910} & \textbf{0.855} & 0.702 & 0.818 & 0.742 & 0.714 & 8.569 & \textbf{9.819} & 6.326 \\ 
\textbf{Baseline + RR SegSE}  & \textbf{0.866} & 0.766 & \textbf{0.698} & 0.898 & 0.820 & \textbf{0.718} & 0.846 & \textbf{0.786} & \textbf{0.718} & \textbf{8.475} & 10.513 & \textbf{6.131} \\ \bottomrule
\end{tabular}}
\label{tab:res_test}
\end{table*}

\begin{figure}[t]
\centering
\includegraphics[width=1.0\columnwidth, keepaspectratio]{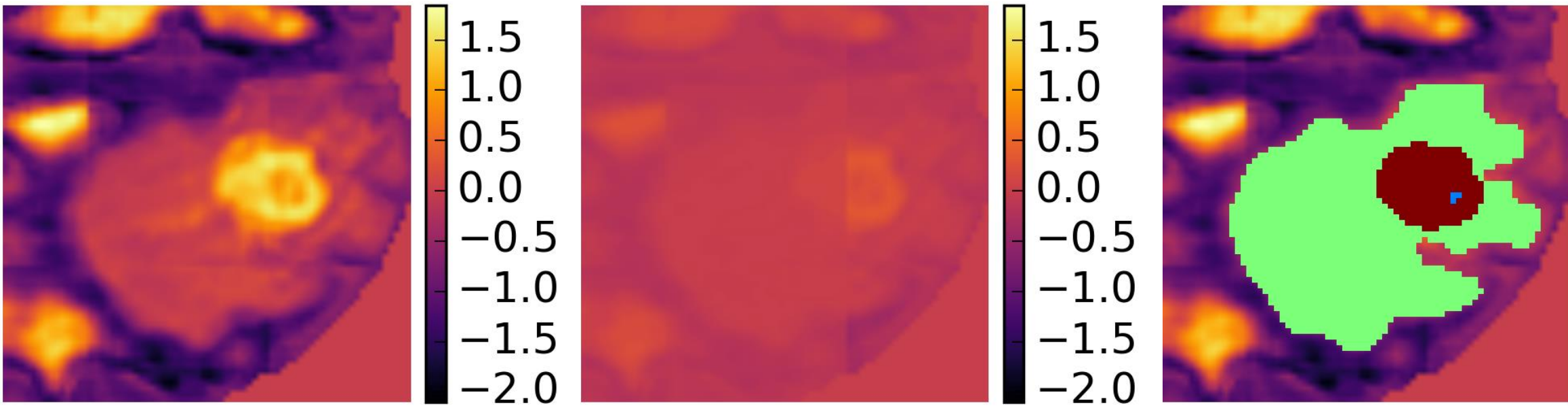}
\caption{Recalibration of a potentially relevant feature map for a small tumor core that is suppressed by the SE block \cite{hu2017squeeze}. From left to right: the feature map before and after recalibration, and the manual segmentation.}
\label{fig:recal_senet}
\end{figure}

In Table \ref{tab:res_test} we can also observe the results obtained by recalibration of the feature maps with recombination. Considering DC, the original SE block \cite{hu2017squeeze} yields worse scores in all tumor classes when compared with the recombination alone. Indeed, it manages to improve over the Baseline only in the tumor core region. The enhancing region is the region that suffers the most, since its DC deteriorates to values even lower than the baseline, which is due to a large decrease in sensitivity. Hence, it may be in accordance with the intuition that the SE block may end up suppressing features that are important for thinner and smaller structures, as observed in Fig. \ref{fig:recal_senet}. This block behaves in this way because it squeezes whole feature maps through their average and recalibrates them as a whole. Since the sensitivity of the enhancing region decreased, we may conclude that it is under-segmenting this region. Moreover, when we observe the results obtained by the different proposed attention mechanisms, DC and Sensitivity of the enhancing region are always better than with the SE block. Note that all the other settings recalibrate the feature maps in a spatially adaptive way. So, we conclude that the SE block, acting as whole feature map recalibration is not adapted for segmentation with FCNs.

Considering the RR block with spatially adaptive recalibration, we may conclude that variants 1 and 2 also achieve worse DC when comparing with the baseline with recombination (Baseline + Recomb.), although in general they perform better in terms of HD$_{95}$. Still, they recover some enhancing tumor that the original SE (Baseline + RR SE) is unable to detect. However, the proposed RR block with SegSE achieves the best results. The DC of the tumor core and the whole tumor are similar to the ones obtained by recombination alone. Nevertheless, we observe an improvement in the enhancing region that achieved the highest score, being the tumor region with the finest details. This was a result of balanced PPV and sensitivity scores that denotes a good quality segmentation, since the HD$_{95}$ was simultaneously the lowest among all the evaluated settings. Additionally, it achieved the best HD$_{95}$ for the whole tumor, and the second for the core region. Qualitatively examining Fig. \ref{fig:seg}, the proposed RR block with SegSE appears to result in better segmentations. The reasons why RR with SegSE performs better than RR with variants 1 and 2 may be because variant 1 does not consider any context, while variant 2 suffers from the checkerboard effect introduced by pooling and up-sampling. 

In Supplementary Materials it can be found results obtained with 3D versions of the baseline networks and the proposed RR with SegSE. We observe that the proposed blocks also improve the performance in a 3D setting.

A drawback of the SegSE block is that it increases the number of parameters. Therefore, to evaluate if its performance is due to the extra capacity, we proportionally increased the width of the baseline, such that the number of parameters becomes similar to the baseline + RR SegSE network. The results obtained with this larger network can be found in Table \ref{tab:res_test} as Wide Baseline. It is possible to observe from DC and HD$_{95}$ that it generally performed worse than the proposed RR SegSE block. Indeed, the Wide Baseline was able to achieve better scores only in the whole tumor and tumor core classes in the PPV metric. So, the improvements of the proposed block are due to learning better features, and not directly to the higher number of parameters.

\begin{figure*}[t]
\centering
\includegraphics[width=0.85\textwidth, keepaspectratio]{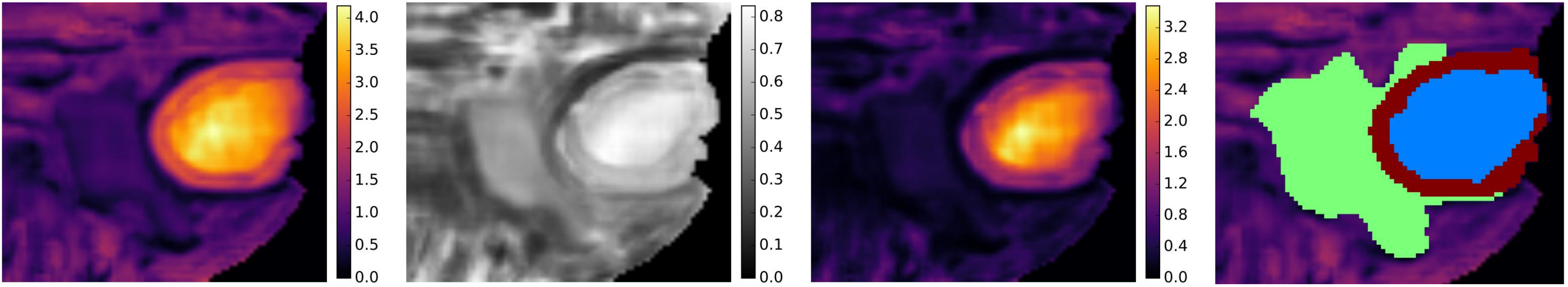}
\caption{Recalibration of a feature map with the SegSE block. From left to right: the feature map before recalibration, the recalibration factor/excitation feature map, the feature map after recalibration, and the manual segmentation.}
\label{fig:recal}
\end{figure*}

To further inspect the effect of the spatially adaptive recalibration procedure with the SegSE block, we can observe a feature map in Fig. \ref{fig:recal}. In the left side it is represented a feature map before recalibration, where we can observe a high response for the tumor core, but also substantial response in the surrounding structures. After recalibration, however, we note how the feature map enhances the tumor core, while suppressing most of the surroundings. Hence, we may conclude that the SegSE block is able to selectively and adaptively recalibrate regions of interest in the feature maps, thus acting as an intra-channel attention mechanism.

\subsection{Brain tumor segmentation}
\label{sec:brats}

Table \ref{tab:res_leaderboard} presents the results obtained in BRATS 2017 Leaderboard with the baseline FCN, the RR with SE block, and the RR with SegSE block\footnote{In the earlier version of this work \cite{segse} the fine-tuning stage described in Subsection \ref{sec:brain_tumor_train} was not performed. Hence, the results in the Leaderboard set reported in \cite{segse} were obtained after training the models with 60\% of the subjects in BRATS 2017 Training set. This explains the improved results in Table \ref{tab:res_leaderboard}.}. It is possible to observe that the FCN with the proposed RR SegSE block outperforms the baseline in both DC and HD$_{95}$ in every tumor regions. This is especially noticeable in the tumor core where the DC improves from 0.758 to 0.798, and the HD$_{95}$ decreases from 11.1 \textit{mm} to 8.947 \textit{mm}. Additionally, with the RR SegSE block the FCN was able to segment the enhancing tumor better, which is a detailed and difficult region \cite{menze2015multimodal}. This can also be observed when comparing with the RR SE block, indicating that the proposed SegSE block is more suited for segmenting detailed regions. Moreover, the proposed SegSE block outperforms the SE block in terms of HD$_{95}$, which suggests that the contours of the obtained segmentations are more detailed and closer to the annotated ones. 

It is also possible to compare with the state of the art in BRATS 2017 Leaderboard in Table \ref{tab:res_leaderboard}, where all the methods are CNN-based. Additionally, we evaluate the block proposed by Roy et al. \cite{roy} in our multi-class FCN. The authors implement an alternative channel and spatial attention mechanism, where a single spatial attention map is inferred for all channels. We evaluate this alternative using the parameters proposed by Roy, but in the same conditions as our RR SegSE block, i.e., after fine-tuning and in a hierarchical approach using the same binary segmentation FCN. We separate single prediction approaches from ensembles\footnote{In single prediction approaches the final segmentation results from the predictions of a model, instead of the combination of several predictions, as in ensembles.}. The reason is that ensembles have a competitive advantage, since it is known that it is a way of significantly improving the performance if the models make different mistakes, as it alleviates the effect of the possible high variance of each of its models \cite{goodfellow2016deep,kamnitsas_ensemble}. In the considered methods, ensembles resulted from training a variety of FCN architectures with different settings \cite{kamnitsas_ensemble}, from training a FCN in each of the MRI planes (axial, coronal, and sagittal) \cite{wang2017automatic,zhao20173d,jungo2017towards}, or from using several models trained previously for \textit{k}-fold cross-validation in the Training set \cite{isensee2017brain}. 

\begin{table}[t]
\caption{Results obtained in BRATS 2017 Leaderboard set. Bold results show the best score for each tumor region among single prediction approaches. Underlined scores are the best among all methods.}

\setlength{\tabcolsep}{0.5em}
\resizebox{1.0\columnwidth}{!}{
\begin{tabular}{clcccccc}

\specialrule{.2em}{.1em}{.1em}
\textbf{} & \textbf{} &  \multicolumn{3}{c}{\textbf{DC}}  & \multicolumn{3}{c}{\textbf{HD$_{95}$}} \\ \cmidrule(l){3 - 5} \cmidrule(l){6 - 8}
\textbf{Approach} & \textbf{Method} & \multicolumn{1}{l}{\textbf{Whole}} & \multicolumn{1}{l}{\textbf{Core}} & \multicolumn{1}{l}{\textbf{Enh.}} & \multicolumn{1}{l}{\textbf{Whole}} & \multicolumn{1}{l}{\textbf{Core}} & \multicolumn{1}{l}{\textbf{Enh.}} \\ \midrule
\multirow{5}{*}{Ensemble} & Kamnitsas et al. \cite{kamnitsas_ensemble} & 0.901 & 0.797 & 0.738 & 4.230 & 6.560 & 4.500 \\ 
 & Wang et al. \cite{wang2017automatic} & \underline{0.905} & \underline{0.838} & \underline{0.786} & \underline{3.890} & \underline{6.479} & \underline{3.282} \\ 
 & Zhao et al. \cite{zhao20173d} & 0.887 & 0.794 & 0.754 & --- & --- & --- \\ 
  & Isensee et al. \cite{isensee2017brain} & 0.896 & 0.797 & 0.732 & 6.970 & 9.480 & 4.550 \\
 & Jungo et al. \cite{jungo2017towards} & 0.901 & 0.790 & 0.749 & 5.409 & 7.487 & 5.379 \\ \midrule
\multirow{6}{*}{Single prediction} & Islam et al. \cite{islam} & 0.876 & 0.761 & 0.689 & 9.820 & 12.361 & 12.938 \\ 
 & Jesson et al. \cite{jesson} & \textbf{0.899} & 0.751 & 0.713 & \textbf{4.160} & \textbf{8.650} & 6.980 \\ 
 &  Roy et al. \cite{roy} & 0.892 & 0.793 & 0.716 & 6.735 & 9.806 & 6.612 \\\cmidrule{2 - 8}
 & Baseline & 0.889 & 0.758 & 0.719 & 6.581 & 11.100 & 5.738 \\ 
 & Baseline + RR SE & 0.891 & \textbf{0.799} & 0.704 & 7.270 & 11.180 & 5.840 \\
 & \textbf{Baseline + RR SegSE} & 0.895 & 0.798 & \textbf{0.733} & 5.920 & 8.947 & \textbf{5.074} \\ \bottomrule
\end{tabular}}
\label{tab:res_leaderboard}
\end{table}

Comparing the single prediction approaches, it is possible to observe that our Baseline is competitive. This is relevant in the sense that it is hard to improve over a competitive approach, which sustains the added value of the RR SegSE block for semantic segmentation. In fact, our FCN with this block outperforms the other single prediction approaches in DC of the enhancing and the tumor core, as well as in HD$_{95}$ of the enhancing tumor, being very close to Jesson et al. \cite{jesson} in DC of the whole tumor. However, Jesson used a FCN with multiple prediction layers and loss functions in different scales. Additionally, the authors employed a learning curriculum to deal with class imbalance. Comparing with Roy et al. \cite{roy}, we observe that their block obtains lower, but competitive, DC of complete and core, but the proposed RR SegSE block enables us to achieve better scores in DC of enhancing tumor and HD$_{95}$ of all tumor regions. The block proposed by Roy includes a branch where the original SE block is used to learn to recalibrate at the channels level. However, according to our observations with the Baseline + RR SE variant in Table \ref{tab:res_test} and Fig. \ref{fig:recal_senet}, it may not be adequate for segmenting fine structures such as enhancing tumor, which may be a reason why the proposed RR SegSE block performs better. When we compare with ensemble methods, we observe that incorporating the RR SegSE block allowed our approach to achieve competitive results, especially in DC. The HD$_{95}$ metric suffers more from the presence of false positive detections, especially if they are far away from the object, and ensembles may effectively tackle this problem since different models do different mistakes. However, our results in DC are similar to the ones obtained by Kamnitsas et al. \cite{kamnitsas_ensemble}, who won BRATS 2017 Challenge edition by generalizing well to the Challenge set, despite the fact that Wang et al. \cite{wang2017automatic} held the best Leaderboard set results.

In Table \ref{tab:res_2013} we present the results on BRATS 2013 Leaderboard and Challenge sets. Even though these datasets are older and smaller, they allow us to compare with a larger variety of methods, such as CNN-based \cite{pereira2016brain,shen,zhao2018deep,havaei2017brain}, Random Forest and handcrafted features-based methods \cite{tustison2015optimal,pinto2018hierarchical}, and multi-atlas patch-based methods \cite{cordier2016patch}. Comparing our Baseline with the FCN with RR SegSE block, we can observe that the latter achieves an overall better performance in terms of DC in both the Leaderboard and Challenge set. Still, it is worth noting the competitiveness of the Baseline FCN. We also observe that the proposed RR SegSE block obtains better Dice and Sensitivity than the RR SE block, but lower PPV, except in the core region in the Challenge set. This may be a hint that the SE block may fall into undersegmentation by suppressing the tumor regions. Comparing with the state of the art, we verify that in the Challenge set our proposal obtains higher DC and Sensitivity than the other approaches. In the case of Leaderboard, the results are competitive with Zhao et al. \cite{zhao2018deep}. However, their approach is based on an ensemble of three CNNs (one for each MRI plane), followed by a Conditional Random Field formulated as a Recurrent Neural Network and extensive post-processing. To our knowledge, these are new state-of-the-art results in BRATS 2013 dataset.

\begin{table*}[!htbp]
\caption{Results obtained in BRATS 2013 Challenge and Leaderboard sets. Bold results show the best score for each tumor region, in each set.}
\centering
\resizebox{0.72\linewidth}{!}{
\begin{tabular}{clcccccccccc}
\specialrule{.2em}{.1em}{.1em} 
 &  & & \multicolumn{3}{c}{\textbf{DC}}  &  \multicolumn{3}{c}{\textbf{PPV}}  & \multicolumn{3}{c}{\textbf{Sensitivity}}  \\ \cmidrule(l){4-6} \cmidrule(l){7-9} \cmidrule(l){10-12}
 & \textbf{Method} & \textbf{Year} & \multicolumn{1}{l}{\textbf{Whole}} & \multicolumn{1}{l}{\textbf{Core}} & \multicolumn{1}{l}{\textbf{Enh.}} & \multicolumn{1}{l}{\textbf{Whole}} & \multicolumn{1}{l}{\textbf{Core}} & \multicolumn{1}{l}{\textbf{Enh.}} & \multicolumn{1}{l}{\textbf{Whole}} & \multicolumn{1}{l}{\textbf{Core}} & \multicolumn{1}{l}{\textbf{Enh.}} \\ \midrule
\multirow{11}{*}{\rotatebox[origin=c]{90}{\textbf{Challenge}}} & Tustison et al. \cite{tustison2015optimal} & 2015  & 0.87 & 0.78 & 0.74 & 0.85 & 0.74 & 0.69 & 0.89 & 0.88 & 0.83 \\
\multicolumn{ 1}{l}{} & Pereira et al. \cite{pereira2016brain} & 2016 & 0.88 & 0.83 & 0.77 & 0.88 & \textbf{0.87} & 0.74 & 0.89 & 0.83 & 0.81 \\ 
\multicolumn{ 1}{l}{} & Shen et al. \cite{shen} & 2017 & 0.88 & 0.83 & 0.76 & 0.87 & \textbf{0.87} & 0.73 & 0.90 & 0.81 & 0.81 \\ 
\multicolumn{ 1}{l}{} & Zhao et al. \cite{zhao2018deep} & 2018 & 0.88 & \textbf{0.84} & 0.77 & 0.90 & \textbf{0.87} & 0.76 & 0.86 & 0.82 & 0.80 \\ 
\multicolumn{ 1}{l}{} & Havaei et al. \cite{havaei2017brain} & 2017 & 0.88 & 0.79 & 0.73 & \textbf{0.89} & 0.79 & 0.68 & 0.87 & 0.79 & 0.80 \\ 
\multicolumn{ 1}{l}{} & Pinto et al. \cite{pinto2018hierarchical} & 2018 & 0.85 & 0.78 & 0.75 & 0.88 & 0.86 & 0.71 & 0.84 & 0.73 & 0.80 \\ 
\multicolumn{ 1}{l}{} & Cordier et al. \cite{cordier2016patch} & 2016 & 0.87 & 0.77 & 0.72 & 0.85 & 0.80 & 0.71 & 0.89 & 0.76 & 0.77 \\ \cmidrule(l){2-12}
\multicolumn{ 1}{l}{} & Baseline & --- & 0.87 & 0.83 & 0.77 & 0.81 & 0.81 & 0.71 & \textbf{0.94} & 0.88 & 0.87 \\ 
\multicolumn{ 1}{l}{} & Baseline + RR SE & --- & 0.88 & 0.81 & 0.76 & 0.88 & 0.83 & \textbf{0.83} & 0.89 & 0.81 & 0.71 \\ 
\multicolumn{ 1}{l}{} & \textbf{Baseline + RR SegSE} & --- & \textbf{0.89} & \textbf{0.84} & \textbf{0.78} & 0.86 & 0.83 & 0.71 & 0.93 & \textbf{0.89} & \textbf{0.88} \\ \specialrule{.2em}{.1em}{.1em} 
\multirow{10}{*}{\rotatebox[origin=c]{90}{\textbf{Leaderboard}}} & Tustison et al. \cite{tustison2015optimal} & 2015 & 0.79 & 0.65 & 0.53 & 0.83 & 0.70 & 0.51 & 0.81 & 0.73 & 0.66 \\ 
\multicolumn{ 1}{l}{} & Pereira et al. \cite{pereira2016brain} & 2016 & 0.84 & 0.72 & 0.62 & 0.85 & 0.82 & 0.60 & 0.86 & 0.76 & 0.68 \\ 
\multicolumn{ 1}{l}{} & Zhao et al. \cite{zhao2018deep} & 2018 & \textbf{0.86} & \textbf{0.73} & 0.62 & 0.89 & 0.77 & 0.60 & 0.85 & 0.77 & 0.69 \\ 
\multicolumn{ 1}{l}{} & Havaei et al. \cite{havaei2017brain} & 2017 & 0.84 & 0.71 & 0.57 & 0.88 & 0.79 & 0.54 & 0.84 & 0.72 & 0.68 \\ 
\multicolumn{ 1}{l}{} & Pinto et al. \cite{pinto2018hierarchical} & 2018 & 0.84 & 0.68 & 0.60 & 0.86 & \textbf{0.86} & 0.61 & 0.84 & 0.65 & 0.68 \\
\multicolumn{ 1}{l}{} & Cordier et al. \cite{cordier2016patch} & 2016 & 0.72 & 0.55 & 0.47 & 0.71 & 0.56 & 0.41 & 0.84 & 0.69 & 0.59 \\ \cmidrule(l){2-12}
\multicolumn{ 1}{l}{} & Baseline & --- & 0.85 & \textbf{0.73} & 0.62 & 0.81 & 0.74 & 0.57 & \textbf{0.93} & \textbf{0.79} & 0.72 \\ 
\multicolumn{ 1}{l}{} & Baseline + RR SE & --- & 0.84 & 0.71 & 0.60 & \textbf{0.90} & 0.80 & \textbf{0.71} & 0.83 & 0.72 & 0.60 \\ 
\multicolumn{ 1}{l}{} & \textbf{Baseline + RR SegSE} & --- & \textbf{0.86} & 0.72 & \textbf{0.63} & 0.84 & 0.78 & 0.61 & 0.92 & 0.76 & \textbf{0.73} \\ \bottomrule
\end{tabular}}
\label{tab:res_2013}
\end{table*}

As previously mentioned, the hierarchical segmentation approach for brain tumor segmentation is helpful when dealing with class imbalance. A possible limitation of this strategy is that if the first stage is poor in a given challenging subject, it will affect the quality of the second, multi-class, stage. Nevertheless, the results in Table \ref{tab:res_leaderboard} and Table \ref{tab:res_2013} are competitive with the state of the art, which suggests that the proposed method does not fail to segment more tumors than the other methods, at least in these datasets. 

In summary, in both BRATS 2017 Leaderboard and BRATS 2013 Leaderboard and Challenge sets we can draw two observations. 1) The variant with the RR SegSE block improves over our competitive Baseline FCN. 2) Although our baseline FCN is simple, our results with the proposed block are competitive, or superior, when comparing with the state of the art.

\subsection{Stroke penumbra estimation}
\label{sec:spes}

Table \ref{tab:res:spes} depicts the results obtained in stroke penumbra estimation in the SPES dataset with the proposed approach, as well as other state-of-the-art methods. We can observe that the Baseline FCN is competitive when comparing with the other methods. Nevertheless, the Baseline + RR SegSE block was able to improve results further in terms of both DC and ASSD. Therefore, the benefits of the proposed RR SegSE block appear to generalize in this problem as well. Moreover, the proposed SegSE block outperforms the SE block, both in DC and ASSD. In fact, the SE has a negative effect over the Baseline FCN in SPES. Although in Table \ref{tab:res:spes} we report the metrics used in \cite{maier2017isles}, the Baseline + RR SE obtained PPV and Sensitivity of $0.86 \pm 0.09$ and $0.77 \pm 0.17$, respectively, while Baseline + RR SegSE scored $0.81 \pm 0.11$ and $0.84 \pm 0.13$ in terms of PPV and Sensitivity, respectively. Hence, we conclude that in SPES the SE block may suppress large portions of the lesions, since it suppresses complete feature maps. In contrast, the SegSE block is spatially adaptive, thus adaptively suppressing regions of the feature maps. In this way, the proposed blocks appear to be able to balance PPV and Sensitivity.

\begin{table}[!b]
\caption{Results obtained in SPES Challenge set. Bold results show the best score considering mean and standard deviation in each metric.}
\centering
\resizebox{0.75\linewidth}{!}{
\begin{tabular}{lcc}
\specialrule{.2em}{.1em}{.1em} 
\textbf{Method} & \textbf{DC} & \textbf{ASSD} \\ \midrule
CH-Insel \cite{maier2017isles} & \textbf{0.82 $\pm$ 0.08} & 1.65 $\pm$ 1.40 \\ 
Cl{\`e}rigues et al. \cite{clerigues2018sunet} & 0.82 $\pm$ 0.09 & --- \\ 
DE-Uzl \cite{maier2017isles} & 0.81 $\pm$ 0.09 & 1.36 $\pm$ 0.74 \\ 
BE-Kul2 \cite{maier2017isles} & 0.78 $\pm$ 0.09 & 2.77 $\pm$ 3.27 \\ 
CN-Neu \cite{maier2017isles} & 0.76 $\pm$ 0.09 & 2.29 $\pm$ 1.76 \\ 
Pereira et al. \cite{pereira2018enhancing} & 0.75 $\pm$ 0.14 & 2.43 $\pm$ 1.93 \\ 
DE-UKF \cite{maier2017isles} & 0.73 $\pm$ 0.13 & 2.44 $\pm$ 1.93 \\ 
BE-Kul1 \cite{maier2017isles} & 0.67 $\pm$ 0.24 & 4.00 $\pm$ 3.39 \\ 
CA-Usher \cite{maier2017isles} & 0.54 $\pm$ 0.26 & 5.53 $\pm$ 7.59 \\ \midrule
Baseline & 0.81 $\pm$ 0.10 & 1.30 $\pm$ 0.74 \\
Baseline + RR SE  & 0.80 $\pm$ 0.12 & 1.54 $\pm$ 1.40 \\ 
\textbf{Baseline + RR SegSE}  & 0.82 $\pm$ 0.09 & \textbf{1.27 $\pm$ 0.72} \\ \bottomrule
\end{tabular}}
\label{tab:res:spes}
\end{table}

Among the compared methods, only CA-Usher \cite{maier2017isles}, Cl{\`e}rigues et al. \cite{clerigues2018sunet}, and Pereira et al. \cite{pereira2018enhancing} are based on Representation Learning approaches. While CA-Usher and Cl{\`e}rigues employ CNNs, Pereira learned features from data using Restricted Boltzmann Machines. Observing Table \ref{tab:res:spes}, we verify that our approach achieves better scores than CA-Usher and Pereira et al. \cite{pereira2018enhancing}, which may be due to those models being shallow. When we compare with the more recent FCN-based approach by Cl{\`e}rigues et al. \cite{clerigues2018sunet}, we verify that both approaches obtained similar DC. The results of the participants in the SPES challenge are officially reported in \cite{maier2017isles} in terms of DC and ASSD. However, Cl{\`e}rigues et al. \cite{clerigues2018sunet} opted to use HD instead of ASSD. In this way, the authors report a HD score of $23.9 \pm 13.5 \ mm$, while the proposed Baseline + RR SegSE obtained $22.67 \pm 11.47 \ mm$. Therefore, although our approach performs similar to Cl{\`e}rigues et al. \cite{clerigues2018sunet} in DC, it is slightly better in HD, which may indicate a better delineation of the lesions. Two other top performing methods among the compared ones are CH-Insel and DE-Uzl, being both based on machine learning approaches with handcrafted features. The method proposed by the CH-Insel team employed the Random Forest classifier with texture features, and a bootstrapping scheme during training at the subject and voxel level. DE-Uzl is also based on the Random Forest classifier, but using intensity, hemispheric difference, local histograms, and center distances as features. Note that while CH-Insel obtained better DC than DE-Uzl, in terms of ASSD it was the other way around. When we compare the proposed Baseline + RR SegSE with those methods, it is possible to observe that it achieves the same DC as CH-Insel and better ASSD than both of those teams. Hence, the proposed approach appears to be competitive or outperform the state of the art in the SPES dataset.

\subsection{Ischemic stroke lesion outcome prediction}
\label{sec:isles}

Table \ref{tab:res_isles} presents the results obtained in ISLES 2017 Challenge set regarding ischemic stroke lesion outcome prediction. We observe that most of the top performing methods obtain higher sensitivity than PPV scores. This may be due to the existence of very small lesions with just a few voxels. Therefore, the models may tend towards over-estimation to avoid missing any lesion. Taking the Baseline FCN into consideration, we note that it is competitive with the state of the art. Indeed, it achieves a DC score of 0.31, placing it only below HKU-1. However, the proposed SegSE block is able to improve the DC score by 0.03, achieving a score of 0.34. This results from an increase of the Sensitivity metric, while PPV decreases only 0.01. Hence, we conclude that the proposed blocks contribute for predicting the lesions better. This is maybe due to their ability to suppress irrelevant regions, thus allowing the model to focus on the important parts of the image. The SegSE block also outperforms the SE block in this application, although the latter is also able to improve the DC of the Baseline FCN to 0.32. The RR SE block results in a large gap between Sensitivity and PPV, which may result in a poorer prediction than the one obtained by the RR SegSE block. Indeed, this is confirmed by the HD and ASSD of $29.09 \pm 14.88 \ mm$ and $5.17 \pm 3.25 \ mm$, respectively, obtained by our RR SegSE block, against the HD and ASSD of $35.58 \pm 15.58 \ mm$ and $6.32 \pm 4.33 \ mm$, respectively, of the RR SE block.

\begin{table}[t]
\caption{Results obtained in ISLES 2017 Challenge set. Bold results show the best score considering mean and standard deviation in each metric. }
\centering
\resizebox{0.85\linewidth}{!}{
\begin{tabular}{lcccc}
\specialrule{.2em}{.1em}{.1em} 
\textbf{Method} & \textbf{DC} & \textbf{PPV} & \textbf{Sensitivity} \\ \midrule
SNU-2 \cite{winzeck2018isles} & 0.31 $\pm$ 0.23 & 0.36 $\pm$ 0.27 & 0.45 $\pm$ 0.31 \\ 
UL \cite{winzeck2018isles} & 0.29 $\pm$ 0.21 & 0.34 $\pm$ 0.26 & 0.51 $\pm$ 0.33 \\ 
HKU-1 \cite{winzeck2018isles} & 0.32 $\pm$ 0.23 & 0.34 $\pm$ 0.27 & 0.39 $\pm$ 0.28 \\ 
INESC \cite{winzeck2018isles} & 0.30 $\pm$ 0.22 & 0.34 $\pm$ 0.27 & 0.51 $\pm$ 0.31 \\ 
KUL \cite{winzeck2018isles} & 0.27 $\pm$ 0.22 & \textbf{0.44 $\pm$ 0.33} & 0.39 $\pm$ 0.31 \\ 
SNU-1 \cite{winzeck2018isles} & 0.28 $\pm$ 0.23 & 0.36 $\pm$ 0.31 & 0.41 $\pm$ 0.31 \\ 
UM \cite{winzeck2018isles} & 0.29 $\pm$ 0.22 & 0.26 $\pm$ 0.24 & 0.61 $\pm$ 0.28 \\ 
MIPT \cite{winzeck2018isles} & 0.27 $\pm$ 0.20 & 0.31 $\pm$ 0.28 & 0.39 $\pm$ 0.29 \\ 
SU \cite{winzeck2018isles} & 0.26 $\pm$ 0.21 & 0.28 $\pm$ 0.25 & 0.56 $\pm$ 0.26 \\ 
KUL \cite{winzeck2018isles} & 0.17 $\pm$ 0.16 & 0.23 $\pm$ 0.28 & 0.36 $\pm$ 0.33 \\ 
Pinto et al. \cite{pinto2018enhancing} & 0.29 $\pm$ 0.21 & 0.23 $\pm$ 0.21 & \textbf{0.66 $\pm$ 0.29} \\ \midrule

Baseline & 0.31 $\pm$ 0.22 & 0.37 $\pm$ 0.29 & 0.49 $\pm$ 0.33 \\ 
Baseline + RR SE & 0.32 $\pm$ 0.22 & 0.30 $\pm$ 0.25 & 0.60 $\pm$ 0.29 \\ 
\textbf{Baseline + RR SegSE} & \textbf{0.34 $\pm$ 0.20} & 0.36 $\pm$ 0.25 & 0.55 $\pm$ 0.30 \\ \bottomrule
\end{tabular}}
\label{tab:res_isles}
\end{table}

The methods in Table \ref{tab:res_isles} represent the top-10 methods that participated in ISLES 2017 challenge, together with Pinto et al. \cite{pinto2018enhancing}, which, to our knowledge, was the first approach to incorporate information from the perfusion Dynamic Susceptibility Contrast-enhanced MRI sequence. All of these methods are based on CNNs, mostly FCN-based. Moreover, SNU-1, SNU-2, HKU-1, and MIPT are ensembles of different FCN architectures and training settings. As expected, these ensemble-based approaches are among the highest ranked ones. SNU-2 and HKU-1 obtained DC of 0.31 and 0.32, respectively. However, the proposed Baseline + RR SegSE was able to outperform both methods by achieving a DC of 0.34. Moreover, the standard deviation was also smaller. Regarding PPV, both our method and SNU-2 performed similarly, with a 0.36 score, while the proposed method obtained sensitivity of 0.55, which is higher than both SNU-2 and HKU-1.

Ischemic stroke lesion outcome prediction differs from segmentation in the sense that we are predicting the final infarct core at a three month follow-up acquisition. In this scenario, the proposed blocks were able to improve an already competitive baseline FCN. This may be due to the capabilities of the proposed blocks to suppress irrelevant regions of the features maps, by taking context into account. To the best of our knowledge, the results obtained in ISLES 2017 are state-of-the-art.

\section{Conclusion}
\label{sec:conclusion}

In summary, channel recalibration of feature maps consists in learning the dependencies among channels, and use it to suppress the less relevant ones. Although desirable, this approach is not well suited for semantic segmentation with FCNs, where several voxels in a patch are segmented simultaneously. In this case, a feature map may contain regions that are relevant for certain voxels, but are less important for others. Also, convolutional layers with $1\times 1$ kernels were used before as bottlenecks in a way to decrease the computational complexity. In this work, we propose to use layers with $1 \times 1$ kernels to recombine features, by an expansion followed by compression of the feature maps. Additionally, we propose a spatially adaptive recalibration block. With this block, we are able to suppress only the less relevant regions of the feature maps, while maintaining the important parts, behaving as an intra-channel attention mechanism. The proposed recalibration block (SegSE) employs dilated convolution for aggregating context. Experimentally, we show that the proposed RR with SegSE block leads to improvements over a competitive baseline. This behavior was observed in all three applications: brain tumor segmentation, stroke penumbra estimation, and ischemic stroke lesion outcome prediction. Our baseline FCN is a simple encoder-decoder FCN, and in this work we aimed at studying spatially adaptive recalibration. However, when we added the RR SegSE block we were able to achieve competitive or state-of-the-art results. Finally, the proposed block is general and it should be possible to add it to other FCN architectures for semantic segmentation.

\section*{Acknowledgments}
S\'ergio Pereira was supported by a scholarship from the Funda\c{c}\~ao para a Ci\^encia e Tecnologia (FCT), Portugal (scholarship number PD/BD/105803/2014). This work is supported by FCT with the reference project UID/EEA/04436/2013, COMPETE 2020 with the code POCI-01-0145-FEDER-006941 and by COMPETE: POCI-01-0145-FEDER-007043 and FCT within the Project Scope:UID/CEC/00319/2013.


\bibliographystyle{temp/IEEEtran}
\bibliography{references}

\end{document}